\documentclass[10pt,conference,final,a4paper]{IEEEtran}
\IEEEoverridecommandlockouts
\usepackage{cite}
\usepackage{amsmath,amssymb,amsfonts}
\usepackage{algorithmic}
\usepackage{graphicx}
\graphicspath{{./figures/}{./}}
\DeclareGraphicsExtensions{.pdf,.png,.jpg,.jpeg,.eps}
\usepackage[caption=false]{subfig}
\usepackage{textcomp}
\usepackage{xcolor}
\usepackage{acronym}
\usepackage{bm}
\usepackage{multirow}
\usepackage{anyfontsize}
\usepackage{hyperref}
\usepackage{scalerel}
\usepackage{tikz}
\usetikzlibrary{svg.path}

\definecolor{orcidlogocol}{HTML}{A6CE39}
\tikzset{
  orcidlogo/.pic={
    \fill[orcidlogocol] svg{M256,128c0,70.7-57.3,128-128,128C57.3,256,0,198.7,0,128C0,57.3,57.3,0,128,0C198.7,0,256,57.3,256,128z};
    \fill[white] svg{M86.3,186.2H70.9V79.1h15.4v48.4V186.2z}
                 svg{M108.9,79.1h41.6c39.6,0,57,28.3,57,53.6c0,27.5-21.5,53.6-56.8,53.6h-41.8V79.1z M124.3,172.4h24.5c34.9,0,42.9-26.5,42.9-39.7c0-21.5-13.7-39.7-43.7-39.7h-23.7V172.4z}
                 svg{M88.7,56.8c0,5.5-4.5,10.1-10.1,10.1c-5.6,0-10.1-4.6-10.1-10.1c0-5.6,4.5-10.1,10.1-10.1C84.2,46.7,88.7,51.3,88.7,56.8z};
  }
}

\newcommand\orcidicon[1]{\href{https://orcid.org/#1}{\mbox{\scalerel*{
\begin{tikzpicture}[yscale=-1,transform shape]
\pic{orcidlogo};
\end{tikzpicture}
}{|}}}}

\def\BibTeX{{\rm B\kern-.05em{\sc i\kern-.025em b}\kern-.08em
    T\kern-.1667em\lower.7ex\hbox{E}\kern-.125emX}}

\acrodef{IoT}{Internet of Things}
\acrodef{NDT}{Non-Destructive Testing}
\acrodef{ML}{Machine Learning}
\acrodef{AI}{Artificial Intelligence}
\acrodef{ANN}{Artificial Neural Network}
\acrodef{SVM}{Support Vector Machine}
\acrodef{UTS}{Ultimate Tensile Strength}
\acrodef{GPR}{Gaussian Process Regression}
\acrodef{GP}{Gaussian Process}
\acrodef{MLR}{Multiple Linear Regression}
\acrodef{LSTM}{Long Short Term Memory}
\acrodef{HTM}{Hierarchical Temporal Memory}
\acrodef{CPS}{Cyber Physical System}
\acrodef{PLS}{Partial Least Squares}
\acrodef{RMSE}{Root-Mean-Square Error}
\acrodef{USL}{Upper Specification Limit}
\acrodef{PCA}{Principal Component Analysis}
\acrodef{GMLVQ}{Generalized Matrix Learning Vector Quantization}

\DeclareMathOperator{\USL}{USL}

\usepackage[absolute,showboxes]{textpos}
\TPMargin{5pt}

\newcommand{\copyrightstatement}{
    \begin{textblock}{0.84}(0.08,0.93)    
         \noindent
         \footnotesize
         \copyright 2022 IEEE. Personal use of this material is permitted.
  Permission from IEEE must be obtained for all other uses, in any current or future
  media, including reprinting/republishing this material for advertising or promotional
  purposes, creating new collective works, for resale or redistribution to servers or
  lists, or reuse of any copyrighted component of this work in other works.
    \end{textblock}
}

\begin{document}
\copyrightstatement

\title{An Industry 4.0 example: real-time quality control for steel-based mass production using Machine Learning on non-invasive sensor data
\thanks{We acknowledge funding through the
\emph{Northern Netherlands Region of Smart Factories (RoSF)} consortium, led by the \emph{Noordelijke Ontwikkelings en Investerings Maatschappij (NOM)}, The Netherlands.}
}


\author{
    \IEEEauthorblockN{Michiel Straat\IEEEauthorrefmark{1}, Kevin Koster\IEEEauthorrefmark{2}, Nick Goet\IEEEauthorrefmark{2}, Kerstin Bunte\IEEEauthorrefmark{1}}
    \IEEEauthorblockA{\IEEEauthorrefmark{1}Bernoulli Institute, University of Groningen,
The Netherlands,
    \{michielstraat, kerstin.bunte\}@gmail.com}
    \IEEEauthorblockA{\IEEEauthorrefmark{2}Philips Personal Health, MG Innovation DTN,
    Drachten, The Netherlands,
    \{kevin.koster, nick.goet\}@philips.com}
}


\maketitle


\begin{abstract}
Insufficient steel quality in mass production can cause extremely costly damage to tooling, production downtimes and low quality products. 
Automatic, fast and cheap strategies to estimate essential material properties for quality control, risk mitigation and the prediction of faults are highly desirable. In this work we analyse a high throughput production line of steel-based products. 
Currently, the material quality is checked using manual destructive testing, which is slow, wasteful and covers only a tiny fraction of the material. 
To achieve complete testing coverage our industrial collaborator developed a contactless, non-invasive, electromagnetic sensor to measure all material during production in real-time.
Our contribution is three-fold: 1) We show in a controlled experiment that the sensor can distinguish steel with deliberately altered properties. 2) During 
several months of production 48 steel coils were fully measured non-invasively and additional destructive tests were conducted on samples taken from them to serve as ground truth.
A linear model is fitted to 
predict from the non-invasive measurements two key material properties (yield strength and tensile strength) that normally have to be obtained by destructive tests. The performance is evaluated in leave-one-coil-out cross-validation. 
3) The resulting model is used to analyse the material properties and the relationship with reported product faults on real production data of approximately 108 km of processed material measured with the non-invasive sensor. The model achieves an excellent performance (F3-score of 0.95) predicting material running out of specifications for the tensile strength.
In a second controlled experiment one coil suspected of material faults was sampled 18 times over its full length and repeated non-invasive as well as destructive testing was performed to analyse the relationship between both measurement types in a situation where also product faults and problems during production are expected to occur.
On this coil the model predictions demonstrate that material properties are indeed out of specification near the point for which the products made from the neighbouring coil exhibited faults during production. 
The combination of model predictions and logged product faults shows that if a significant percentage of estimated yield stress values is out of specification, the risk 
of product faults is high. Our analysis demonstrates promising directions for real-time quality control, risk monitoring and fault detection.



\end{abstract}

\begin{IEEEkeywords}
Non-destructive testing, Industry 4.0, Machine Learning, Smart Industry, Soft Sensor, Predictive Models, Industrial Process Monitoring.
\end{IEEEkeywords}

\acresetall

\section{Introduction}
The terms ``Smart Industry'', ``Industry 4.0'', or ``Fourth Industrial Revolution'' \cite{schwab2017fourth} have been coined to describe a vision that includes a wide range of emerging technologies that, when used collaboratively, have the potential to contribute to highly optimized production processes \cite{Lasi2014,Vaidya2018}. 
Several fields are of central importance in the development of the so-called \textit{Smart Factory}, including 
sensor technology, Cyber Physical Systems, the Internet of Things, advanced communication technology, big data analytics, \ac{ML}, \ac{AI} and cloud computing \cite{Chen2017,Castelo2019}. 
Various examples exist in which the successful implementation of these technologies results in higher efficiency, better human decision making and less waste \cite{Vaidya2018}. 
Realizing the large potential of Smart Industry has been recognized as a key factor by governments and industries in ensuring economic competitiveness and sustainability in the next decades.

The main machinery in the production line that we study is a high-speed stamping press that is able to operate on the strip steel at a frequency of 180 strokes per minute. 
It is crucial that all strip steel that enters the production process is of sufficient quality. 
Insufficient material quality results in poor quality of the final product or expensive damage to the production machinery and corresponding production downtimes.
The current quality requirements for the material are given as upper limits of stress [MPa] of the material properties, the so-called \ac{USL}.
Currently, the material quality is checked by performing destructive tests on samples of the steel. 
By means of a tensile test on the sample, key material properties such as yield strength and tensile strength are measured. Although the material properties of the sampled steel can be measured reliably using these methods, the process is manual, slow, produces material waste and is only possible to be performed on a tiny fraction.
Furthermore, it is not a solution for detecting changes in 
material properties over the full length of a steel coil, because more frequent sampling slows down production. 
Destructive tests are therefore not suitable for continuous quality control and
detecting highly local changes in material properties.

In implementations of \textit{soft sensors}, easily obtainable process variables are measured inline, which are converted using statistical or machine learning models to quantities that otherwise have to be measured in expensive, time-consuming lab tests \cite{Jiang2021}. 
An important component of soft sensing in smart industry is \ac{NDT} \cite{Sophian2020}. 
In the steel-based manufacturing industry, \ac{NDT} sensors perform contactless and non-destructive measurements on the steel in real-time and can therefore be used in a high throughput production line to measure all strip steel that enters the process \cite{Garcia2011}. By combining the real-time stream of measurements with appropriate machine learning models, advanced online fault detection and quality control systems can be developed. For instance, in settings where temporal patterns are relevant, Long Short-Term Memory and Gaussian Processes have been used \cite{Malhotra2015,Berns2020}, which can be too computationally expensive for a high-throughput production line. Latent variable models have also been used in industrial settings, such as supervised factor analysis \cite{Zhiqiang2015} and partial least squares \cite{Rosipal2005}. A successful implementation of a real-time quality control system leads to fewer defects in products, improved quality, less production downtime and less material waste. Furthermore, the real-time model estimation of material quality from the inline measurements can be used in the active control of production parameters, which adapts the machinery settings to be optimal for the current specifics of the material \cite{Heingartner2010,Zhiqiang2015,Jiang2021}. 



In this contribution, we develop a real-time quality control and fault detection solution for the high-throughput production line. The measurements are performed at the start of the production line and exactly at the location where the press operates further down the line. Our contributions are three-fold:
1) A model is developed for estimating material properties in real-time from the inline contactless sensor measurements. We use the ground truth material properties of several production coils to fit the model. 2) The model is used for the early detection of insufficient material quality. We show a case where the model estimation of the material properties can detect faulty material in order to prevent production faults. This is shown on a coil from a faulty batch of coils that had already caused product faults. 3) We study the model estimations on 108 km of processed steel and we link the model estimations to reported product faults that occurred during production. These product faults are caused by a crack arising in the product while in the press and it is hypothesized that insufficient material quality is one of the causes.

The paper is organized as follows: in Sec.~\ref{sec:Data} the relevant details of the new industrial datasets are introduced. Subsequently, in Sec.~\ref{sec:Methods} the methods used for the analysis of the data and for the estimation of material properties are discussed. In Sec.~\ref{sec:Results} we present the results of our experiments and discuss them in Sec.~\ref{sec:Discussion}. Lastly, the work is summarized in Sec.~\ref{sec:Conclusion} and an outlook for future work is provided.

\section{Data description and analysis} \label{sec:Data}
The production coils of small strip steel are associated with a ``Heat" number, which identifies the specific elements used in the steel production batch. 
The \ac{NDT} sensor measurements are based on Eddy Currents and the reader is referred to \cite{Garcia2011} for details about these sensors. Here, the measurements are performed at 10 testing frequencies and are denoted by $\bm{x}_i \in \mathbb{R}^{20}$. The first half of the components contain the amplitude gains of each frequency and the second half the phase shifts. Hence, the amplitude gain and phase shift of measurement frequency $j$ are in $x_{ij}$ and $x_{i|j+10}$, respectively. We will also call individual sensor variables ``SV $i$".

\subsection{Controlled experiment: measuring modified steel samples 
} \label{sec:Data:LabModified}
In order to establish an expected lower and upper bound for each sensor variable, steel created with extreme material properties was measured with the contactless sensor and compared to the reference steel.
A selection of nine steel strips was divided
into three groups of three. 
The first group was modified to be ``harder'' and the second group to be ``softer'', i.e. towards larger and lower values of yield- and tensile strength, respectively. The remaining group was left unmodified to serve as reference material.
Twohundred sensor measurements were taken at the start, middle and end of the strips.
We normalized the original sensor measurements $\bm{x}_i \in \mathbb{R}^{20}$ as follows:
\begin{equation} \label{eq:normalize_sensor}
	\bm{x}_i := \frac{\bm{x}_{i} - \bm{P}_{10\%}^H}{\bm{P}_{90\%}^S - \bm{P}_{10\%}^H}\, , \quad
	\bm{x}_i := \bm{x}_i - \bm{\mu}^R\, .
\end{equation}
where $\bm{P}_{10\%}^H \in \mathbb{R}^{20}$ are the 10th percentiles of the values within the \textbf{H}ard group and
$\bm{P}_{90\%}^S \in \mathbb{R}^{20}$ are the 90th percentiles of the values within the \textbf{S}oft group. The vector $\bm{\mu}^R \in \mathbb{R}^{20}$ contains the means of the reference steel samples after the first transformation.
As the hard strips had lower measurement values, the transformation \eqref{eq:normalize_sensor} is effectively min-max normalization, with percentiles 
being estimates of the min and max.
The shift moves the reference material close to zero.

\begin{figure}
\centering
\includegraphics[width=0.65\linewidth]{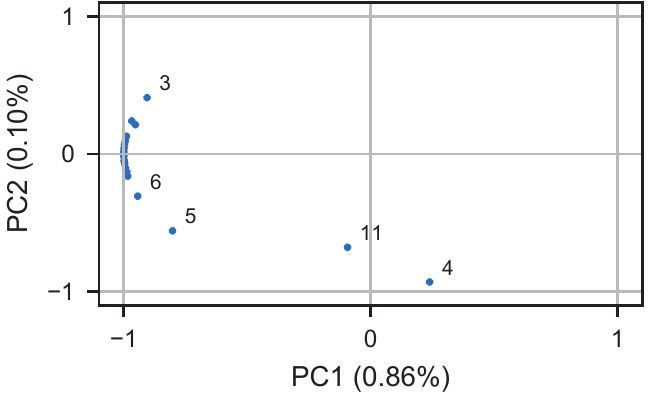}
\caption{Loadings of the first two \ac{PCA} components computed on the standardized hard and soft measurements. Only dissimilar variables are labeled.}
\label{fig:HardZacht_PCA_loadings}
\end{figure}
Fig.~\ref{fig:HardZacht_PCA_loadings} shows the loadings of the first two principal components computed on the standardized measurement data. 
Due to large mutual positive correlations between the variables, the first principal component has large loadings on the majority of the variables and explains already 86\% of the variance in the data. 
Combined with the second principal component, which is loaded mostly on SV 11 and SV 4, 96\% of the variance is explained. 
\begin{figure}
\centerline{\includegraphics[width=0.65\linewidth]{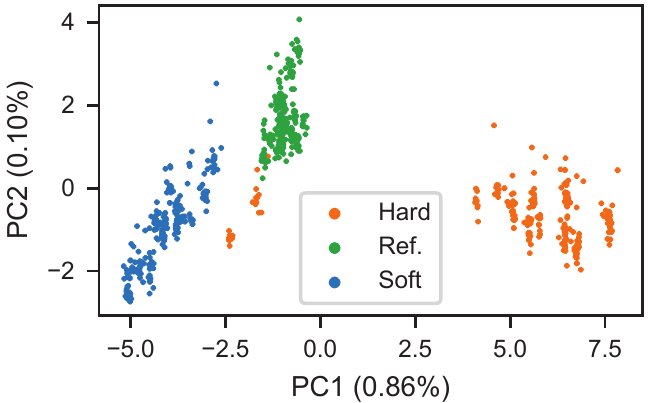}}
\caption{Sensor measurements on steel with different material properties projected on the the first two \ac{PCA} components of the standardized data. 15\% of the total number of measurements is shown, uniform randomly chosen.}
\label{fig:HardZacht_PCA_scores}
\end{figure}
Fig.~\ref{fig:HardZacht_PCA_scores} shows the projection of all data points on the first two principal components. 
In general, the first principal component scores separate the different material properties well.
Remarkably, among the points labeled as hard material, there are two outlier groups of measurements. 
A tensile test of the corresponding strip revealed that the strip had similar material properties to the reference material and therefore this likely indicates a failure in the modification of this strip.



\subsection{Production setting: continuous measurements in the line
}

In this section we discuss the dataset obtained to relate the non-invasive 20 dimensional sensor measurements to material properties obtained by destructive testing.

\subsubsection{Sensor data during production
}
The sensor was installed at the start of the production line to continuously measure the production steel coils. 
This produced a stream of measurements $\bm{x}_i \in \mathbb{R}^{20}$ with a timestamp and the current steel coil identification. 
From each coil a variable number of products is made, thus the range of measurements varies from a few hundred to tens of thousands of products.
In some instances a production stop caused the sensor to produce physically impossible values or no values at all. 
These faulty measurements were removed from the dataset.

\subsubsection{Destructive Tensile tests}
From 47 selected coils a sample at the start of the coil was taken to measure material properties with destructive testing. 
Three tensile tests were performed on each sample to measure yield strength and tensile strength, in the following denoted by ``t1'' and ``t2''.
During the testing period 
one production coil resulted in many products with cracks, hypothesized to be caused by 
insufficient steel quality. 
Hence, it was decided that a related coil from the same heat should be rejected for production and instead be fully measured by the non-invasive sensor as well as frequently sampled for tensile testing.
%
In total, at nine locations distributed over the full length of this coil two samples were taken for tensile testing. 
We label this particular coil as ``Testcoil'' to distinguish it from the rest of the 47 production coils.
\begin{figure}
\centering
\includegraphics[width=0.85\linewidth]{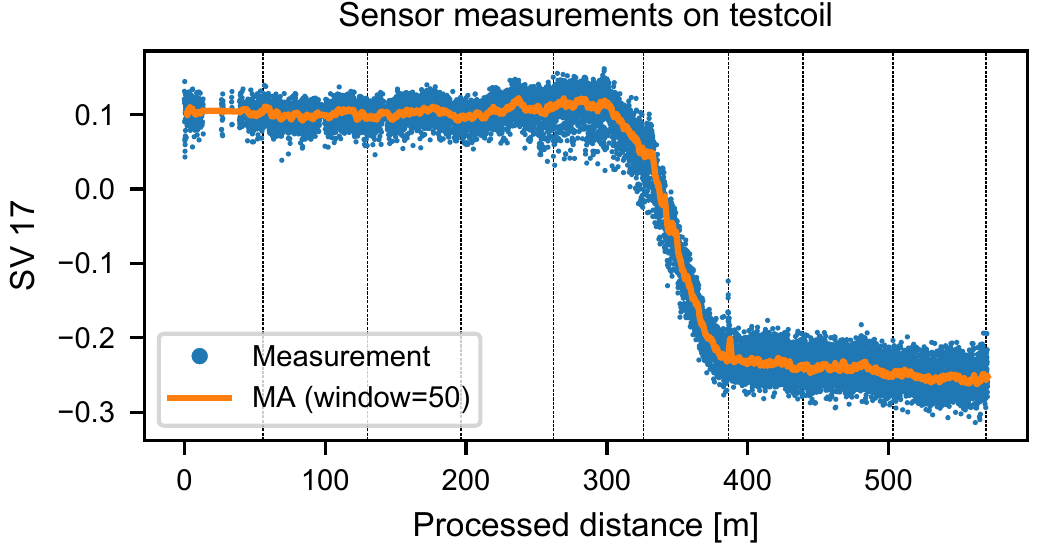}
\caption{\textit{Blue points}: sensor variable 17 measurements made on the testcoil. \textit{Solid orange line}: moving average over 50 measurements. \textit{Dashed black lines}:  locations of the destructive test samples.}
\label{fig:testCoil_sP7}
\end{figure}
Fig.~\ref{fig:testCoil_sP7} shows the value of sensor variable 17 over the full length of this coil, along with the nine locations of the samples at which two tensile tests were performed.

Instances of products that contained cracks were logged during the months of the experiment.
In 17 cases the identification code could be logged of the corresponding inline measured material.
In 25 cases of product faults that occurred over six production coils the hour during which a crack had occurred was logged.
We normalized the sensor measurements using Eq.~\eqref{eq:normalize_sensor}, such that 
values close to zero indicate 
measurements 
similar to the reference material of the experiment in Sec.~\ref{sec:Data:LabModified}. 
Furthermore, negative values are closer to the measurements of the hard material while positive values are closer to the measurements of the soft material.

\begin{figure}
\centerline{\includegraphics[width=0.65\linewidth]{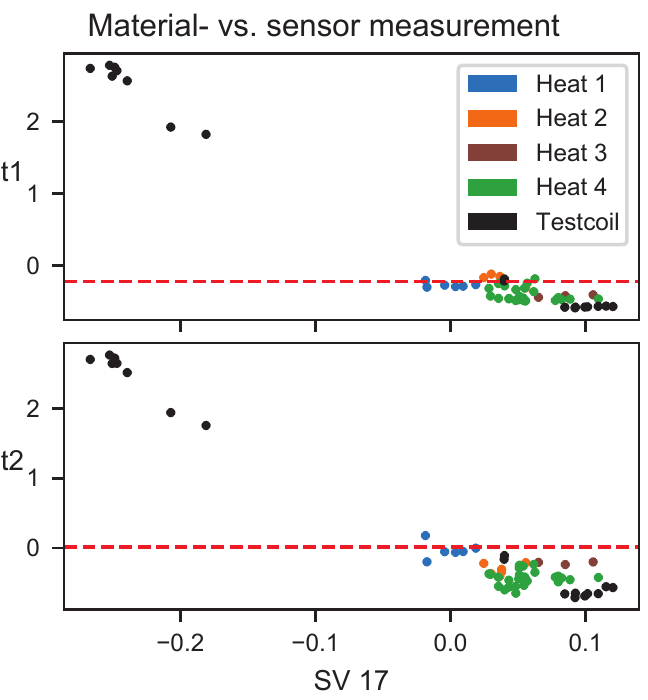}}
\caption{Material properties t1 and t2 against sensor variable SV 17 for the 42 production coils and the testcoil.
Values for t1 and t2 denote the mean of three tensile tests. Values of SV 17 denote the mean
of the first 200 \ac{NDT} sensor measurements for the production coils and for the testcoil the mean around the 18 samples.
Standard deviations are on average about two times the size of the markers.
The dashed red line denotes the \ac{USL} of the respective material properties.}
\label{fig:t1t2vssP7}
\end{figure}

We standardized both material properties t1 and t2 that were obtained from the tensile tests on the 48 coils. 
To relate the tensile tests to the sensor measurements, the mean and standard deviation were computed from the first 200 sensor measurements on the 47 production coils.
Coils with less than 200 measurements were dropped from the data, leaving 42 production coils. 
For the 18 tensile tests performed on the nine samples spanning the full length of the testcoil we computed the mean and standard deviation of the five sensor measurements in the direct neighbourhood.
Fig.~\ref{fig:t1t2vssP7} shows the resulting values of the destructively tested material properties against non-invasive sensor variable 17 for the 43 coils (42 production coils + testcoil, corresponding to 42 tensile tests + 18 tensile tests). 
The \ac{USL} of the material properties is marked in both figures. As can be seen, several points measured on the testcoil had material properties far exceeding the \ac{USL}. 
The corresponding values for SV 17 were also very different from the rest. Some production coils slightly exceeded the \ac{USL} too. 
We observe a negative linear correlation between material properties and sensor measurements. 
In general, coils from the same heat exhibited
similar material properties and
sensor measurements.

\begin{figure}
\centerline{\includegraphics[width=0.65\linewidth]{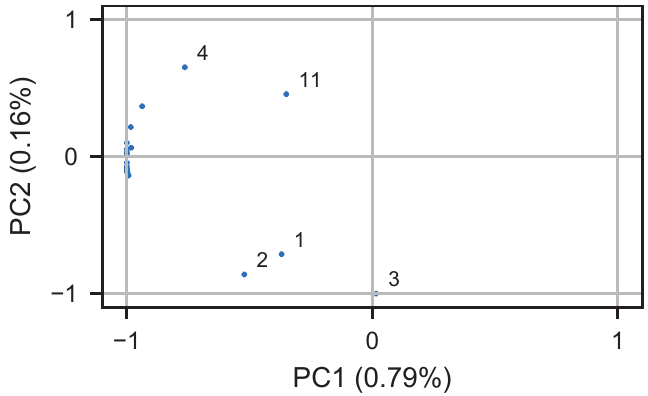}}
\caption{\ac{PCA} loadings of the \ac{NDT} sensor variables computed on the standardized production coil dataset. Only dissimilar variables are labeled.}
\label{fig:productioncoils_PCA_loadings}
\end{figure}
\begin{figure}
\centering
\includegraphics[scale=0.75]{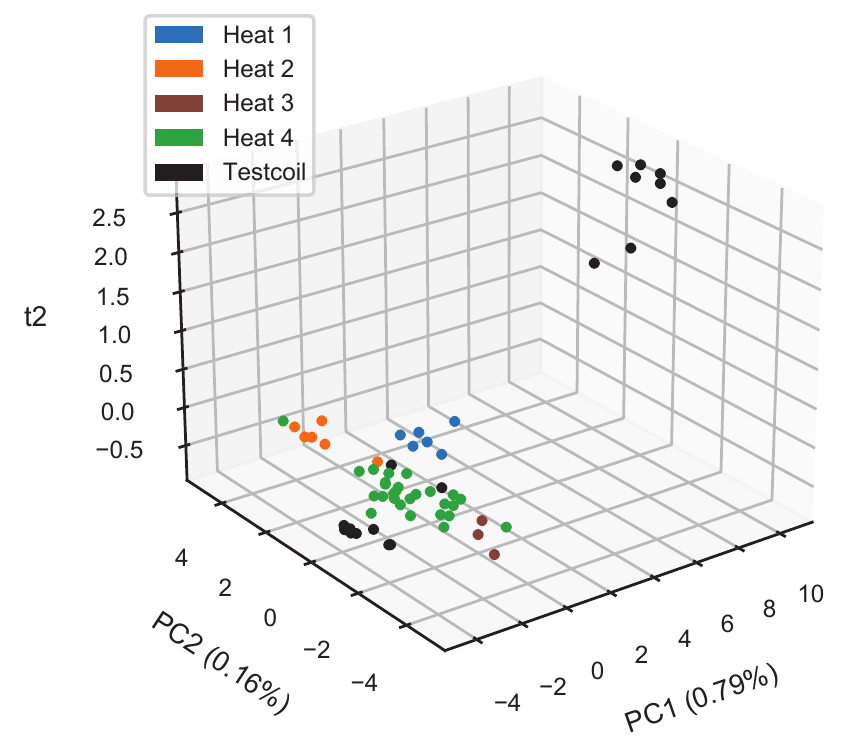}
\caption{The production coil dataset: Material property t2 against the scores on the first two principal components computed by \ac{PCA} on the standardized \ac{NDT} sensor measurements.}
\label{fig:productioncoils_PCA_vs_Rm}
\end{figure}

Fig.~\ref{fig:productioncoils_PCA_loadings} shows the loadings of the first two principal components on the sensor variables, obtained from \ac{PCA} on the full 20-dimensional sensor measurements. 
In general, the loadings and the variation explained by the principal components are similar to those of the controlled experiment of Fig.~\ref{fig:HardZacht_PCA_loadings}.
In Fig.~\ref{fig:productioncoils_PCA_vs_Rm}, the values of material property t2 obtained from the tensile tests on the samples are shown against the projections of the corresponding sensor measurements on the first two principal components.
Note that the points with outlier t2 measurements are only separated from the rest of the sensor measurements on the first principal component. From the scores on the second principal component the different material properties cannot be distinguished.
As seen in Fig.~\ref{fig:testCoil_sP7} the signal is characterized by a band of values and for the test coil it exhibited a large transition in the middle of the coil.
Because of the high redundancy of the variables we ranked the quality of the individual variables by estimating the measurement noise. 
The standard deviation between sample number 2000 and 4000 was computed and divided by the total transition difference of the signal, i.e. for each variable the difference between the first and last value of the moving average of Fig.~\ref{fig:testCoil_sP7}. 
\begin{figure}
\centerline{\includegraphics[width=0.8\linewidth]{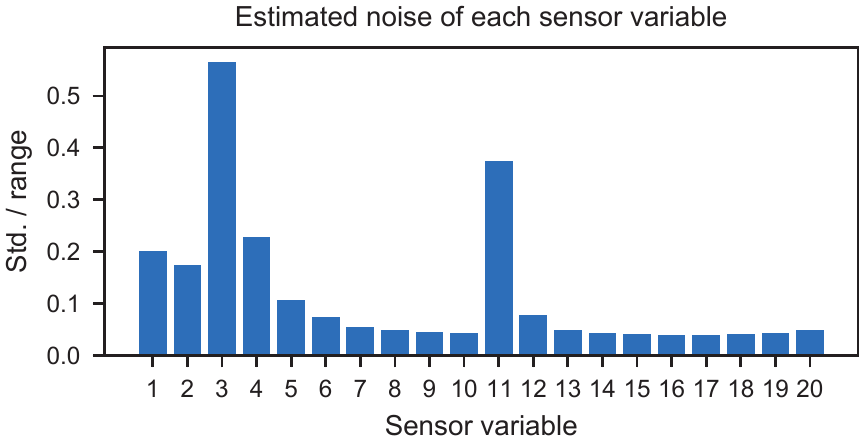}}
\caption{The fraction of the standard deviation with respect to the transition difference in Fig.~\ref{fig:testCoil_sP7}, as an estimation of the measurement noise.}
\label{fig:sensor_noise}
\end{figure}
The value of this fraction is shown in Fig.~\ref{fig:sensor_noise} for each variable. SV 17 had one of the lowest estimated noise values, while SV 3, 4 and 11 had high estimated noise values. Hence, as the second principal component was significantly loaded onto these variables in both datasets, the variance explained by this component was mainly measurement noise.
\begin{table}
\caption{Correlation matrix: Principal Components and material properties without 
(left) and including testcoil points (right)}
\begin{center}
\begin{tabular}{|c|c|c|}
\hline
& \textbf{t1} & \textbf{t2} \\
\hline
\textbf{PC1} & 0.31 & 0.42 \\
\hline
\textbf{PC2}  & 0.45 & 0.04 \\
\hline
\textbf{t1} & 1.00 & 0.38 \\
\hline
\end{tabular}
\quad
\begin{tabular}{|c|c|c|}
\hline
& \textbf{t1} & \textbf{t2} \\
\hline
\textbf{PC1} & 0.97 & 0.97 \\
\hline
\textbf{PC2}  & 0.03 & -0.01 \\
\hline
\textbf{t1} & 1.00 & 0.99 \\
\hline
\end{tabular}
\label{tab:correlationPC_MP}
\end{center}
\end{table}
Table~\ref{tab:correlationPC_MP} contains the Pearson correlation for this dataset computed with and without the testcoil points. Excluding the testcoil points, the correlation with the principal components was much smaller, but still significant.

\section{Methods} 
\label{sec:Methods}
The change in material properties is not considered to result from periodic time variations, but rather local fluctuations in the production of the steel. The analysis in the previous sections demonstrates linear correlations and relationships in our datasets. Therefore a linear model, \ac{PLS}, is considered
for estimating the material quality and fault detection.

For \ac{PLS} regression it is assumed that the data is generated by a smaller number of latent variables than the number of observed variables. Let $n$ be the number of data points, $m$ the number of observed variables and $o$ the number of target variables.
Then for the predictor matrix $\bm{X} \in \mathbb{R}^{n \times m}$, target matrix $\bm{Y} \in \mathbb{R}^{n \times o}$ and assuming $k$ number of latent variables, the \ac{PLS} assumption can be written as follows \cite{Rosipal2005}:
\begin{align}
\begin{split}
\bm{X} = \bm{T} \bm{P}^T + \bm{E}\enspace , \\
\bm{Y} = \bm{U} \bm{Q}^T + \bm{F}\enspace ,
\end{split}
\end{align}
where $\bm{T} \in \mathbb{R}^{n \times k}$ and  $\bm{U} \in \mathbb{R}^{n \times k}$ are the score matrices containing the scores on the $k$ latent variables for each datapoint's input and target, respectively. 
The matrix $\bm{P} \in \mathbb{R}^{m \times k}$ contains the original input variable loadings on the $k$ latent input variables and the matrix $\bm{Q} \in \mathbb{R}^{o \times k}$  contains the original target variable loadings on the $k$ latent target variables.
Lastly, $\bm{E} \in \mathbb{R}^{n \times m}$ and $\bm{F} \in \mathbb{R}^{n \times o}$ are the residuals.
The optimization procedure finds the $k$ latent variables in $\bm{X}$ and $\bm{Y}$ that have maximal covariance.
We used the implementation of \cite{scikit-learn} using the default optimization parameters.
The sensor measurements $\bm{x}_i \in \mathbb{R}^{20}$ are used as inputs and the material properties $\bm{y}_i \in \mathbb{R}^2$ as targets.
We varied the number of latent variables $k$ in cross-validation to determine the optimal value. We evaluated the accuracy of the model using \ac{RMSE} in cross-validation. In each fold of the cross-validation, one coil was left out for validation and the model was fitted on the rest of the coils. This is similar to Leave One Out cross-validation, with the exception of one fold that had 18 datapoints of the testcoil.

Furthermore, we evaluated the accuracy of a binary classifier that was based on thresholding of the estimated material properties $\hat{y}_i \in \mathbb{R}^2$ using the \ac{USL}. We considered the following three classification rules for classifying a measurement as a material fault:
\begin{align} \label{eq:fault_classification_rules}
\begin{split}
\hat{y}_{i1} > \USL(t1)\,, \quad \hat{y}_{i2} > \USL(t2)\,, \\
(\hat{y}_{i1} > \USL(t1)) \, \, | \, \, (\hat{y}_{i2} > \USL(t2))\, .
\end{split}
\end{align}
Hence, material is classified as faulty based on the yield strength and tensile strength individually or based on the combination.
For each rule, it is possible to compute the precision and recall from the resulting classifications and the true labels. We also computed $F_1$ and $F_3$ scores. The $F_3$ score assigns three times more importance to recall over precision, which is more appropriate in our case: a missed material fault means that out of specification material goes into production, which may cause extremely costly damage to the machinery or low quality products. In contrast, a false alarm results in minor cost in the form of wasted material that would have been suitable for production and potentially minor production delays when the material is removed from the coil.

\section{Results} \label{sec:Results}

\subsection{Dataset/Production coils} \label{sec:Results:ProductionCoils}
\begin{figure}
\centerline{\includegraphics[width=0.8\linewidth]{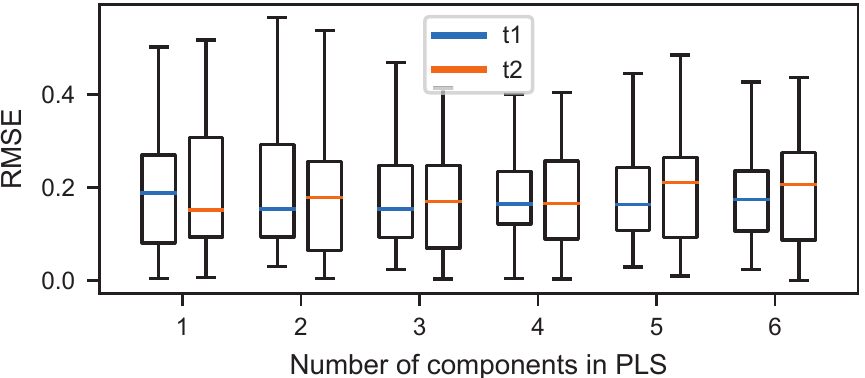}}
\caption{\ac{RMSE} computed as the mean of the \ac{RMSE} obtained on the validation sets in leave-one-coil-out cross-validation vs. the number of components/latent variables in \ac{PLS}.
}
\label{fig:PLS_avgRMSE_K}
\end{figure}
Fig.~\ref{fig:PLS_avgRMSE_K} shows the average \ac{RMSE} obtained by the \ac{PLS} model in 
the one-coil-out cross-validation for increasing number of latent variables $k$. 
Upon further inspection we note that excluding the testcoil points results in
by far the highest \ac{RMSE}, which is an outlier that is not shown in Fig.~\ref{fig:PLS_avgRMSE_K}. 
Thus, one needs to ensure that the full range of variation that is potentially seen in production is included in model fitting, which might require deliberate creation of undesirable material.
Furthermore, it can be seen that the \ac{RMSE} does not decrease significantly by introducing more than one component. 
Hence, \ac{PLS} optimization determined 
one component of the sensor measurements $\bm{X}$ and one component of the material properties $\bm{Y}$ which, due to significant covariance, could be exploited in the regression. 

\begin{figure}
\centering
\includegraphics[width=0.8\linewidth]{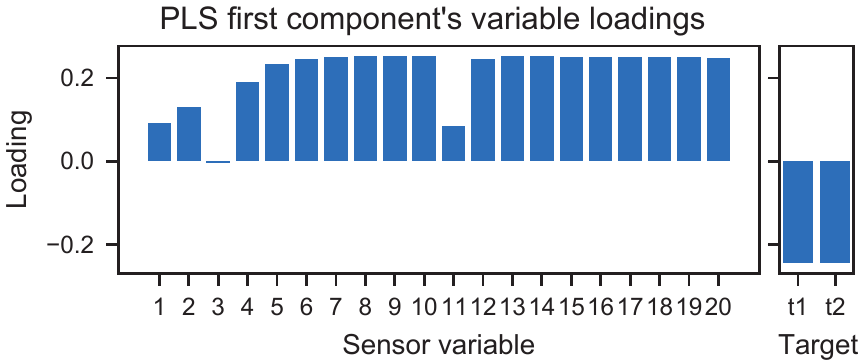}
\caption{Loadings on the component extracted by \ac{PLS} 
of the sensor variables in $\bm{X}$ (\textit{Left}) and 
the destructively tested material properties in $\bm{Y}$ (\textit{Right}).}
\label{fig:PLS_loadingplot}
\end{figure}
Fig.~\ref{fig:PLS_loadingplot} shows the loadings of the first \ac{PLS} component on the variables, for both the sensor variables $\bm{X}$ and the material properties $\bm{Y}$. These loadings were obtained from a \ac{PLS} fit on the entire dataset (42+18 points). As can be seen, the sensor variables 5 to 10 and 12 to 20 had nearly identical loadings on the first component. These loadings were highly similar to the first principal component from the \ac{PCA} of Sec.~\ref{sec:Data}.
The component extracted from $\bm{Y}$ had equal loadings for both material properties.
Since the non-invasive sensor measurements are strongly correlated and one PLS component is sufficient for the task, the question arises if similar performance can be achieved by individual variables.

\begin{figure*}
\centering
\includegraphics[width=0.9\textwidth]{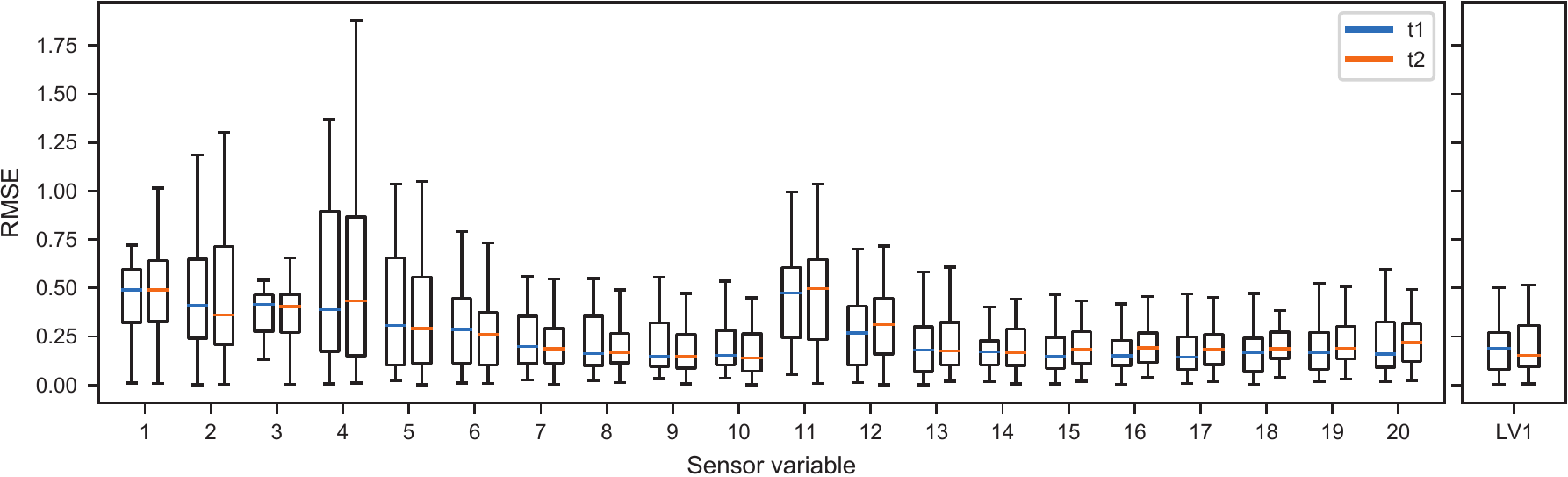}
\caption{\textit{Left}: Cross-validation \ac{RMSE} of linear regression for each sensor variable as predictor of the material properties t1 and t2. \textit{Right}: Cross-validation \ac{RMSE} of PLS with number of components $k=1$. Outliers are not shown.}
\label{fig:linearRegression_RMSE}
\end{figure*}
Fig.~\ref{fig:linearRegression_RMSE} shows the cross-validation \ac{RMSE} for linear regressions with the individual sensor variables as predictor along with the \ac{RMSE} obtained from \ac{PLS}. Linear regressions using one of the higher loaded variables from Fig.~\ref{fig:PLS_loadingplot} had similar performance as the \ac{PLS} model. Although differences were small, the predictions of property t1 and t2 were most accurate when based on SV 17 and SV 10, respectively. 
These sensor variables had low estimated measurement noise in accordance with findings
in Sec.~\ref{sec:Data}, Fig.~\ref{fig:sensor_noise}.
We continued with the \ac{PLS} model, as the latent variable is more robust against sudden changes of the noise pattern in the variables.

\begin{figure}
\centering
\includegraphics[width=0.9\linewidth]{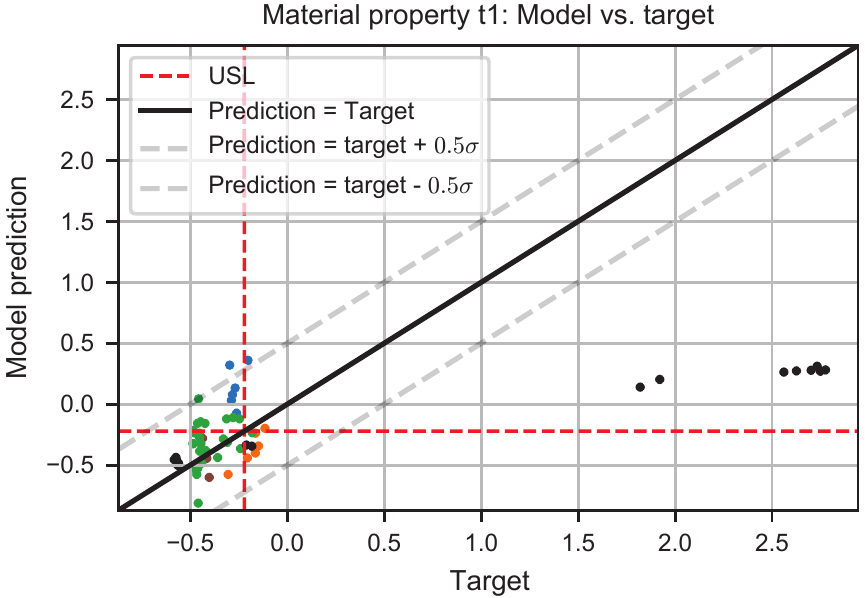}
\\
\includegraphics[width=0.9\linewidth]{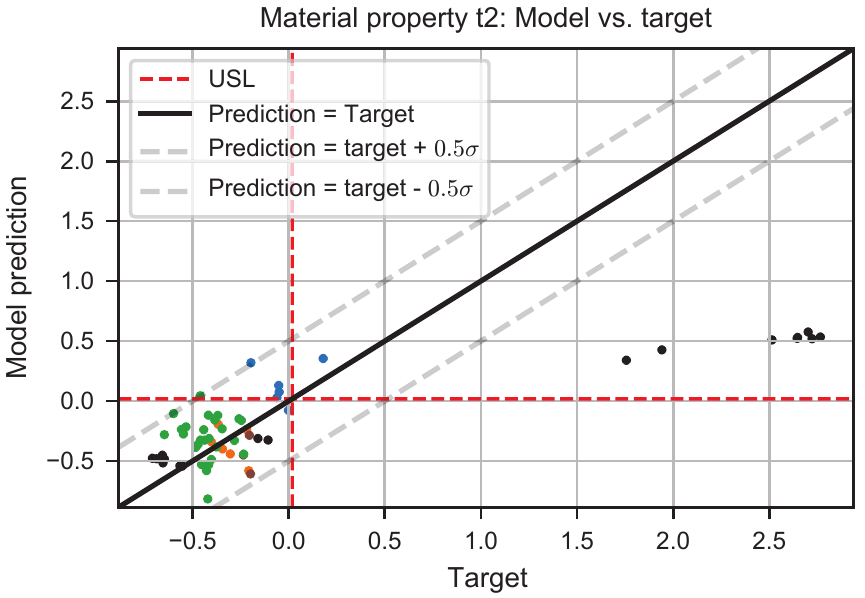}
\caption{One-coil-out cross-validation prediction results of material properties t1 (top panel) and t2 (bottom panel) using the \ac{PLS} model with number of components $k=1$.}
\label{fig:PLS_productionCoils_t2}
\end{figure}
Some coils introduce large variations in the material properties contained in the dataset and the predictions are negatively affected if the full range is not observed.
Fig.~\ref{fig:PLS_productionCoils_t2} shows for both material properties t1 (top) and t2 (bottom) the \ac{PLS} predictions made in the one-coil-out cross-validation against the target output.
Most predictions are within $0.5 \sigma$ of the target variables. The points from the test coil in the validation set are clearly underestimated.
However, the \ac{USL} divides the space into quadrants that are still mostly correctly predicted despite the extreme setting: 
The bottom-left quadrant corresponds to true negative (TN) out-of-specification classifications, the bottom-right to false negatives (FN), top-right to true positive (TP) and top-left to false positive (FP) fault classifications.
\begin{table}
\caption{Performance fault classification based on \ac{PLS} predictions}
\begin{center}
\resizebox{\linewidth}{!}{
	\begin{tabular}{|c|c|c|c|c|c|c|c|c|}
	\hline
	& \textbf{TP} & \textbf{FN} & \textbf{FP} & \textbf{TN} &\textbf{Precision}  & \textbf{Recall} & \textbf{$F_1$} & \textbf{$F_3$}  \\
	\hline
	\textbf{Based on t1} & 10 & 7 & 13 & 30 & 0.43 & 0.59 & 0.50 & 0.57 \\
	\textbf{Based on t2}  & 9 & 0 & 5 & 46 & 0.64 & 1.00 & 0.78 & 0.95 \\
	\textbf{t1 and t2} & 10 & 7 & 13 & 30 & 0.43 & 0.59 & 0.50 & 0.57 \\
	\hline
	\end{tabular}
}
\label{tab:fault_class_PLSresults}
\end{center}
\end{table}
For the classification of the datapoints according to the three classification rules in Eq.~\eqref{eq:fault_classification_rules}, the resulting quantities are listed in Table~\ref{tab:fault_class_PLSresults}, along with the precision, recall, $F_1$-score and $F_3$-score. 
The recall of the fault classification based on t1 and the combination of t1 and t2 was only 0.59 and the corresponding $F_3$-score was 0.57.
The results were identical for these classification rules, because upon further inspection a violation of t2 was always accompanied by a violation of t1, but not vice versa. Hence, we hypothesize that the current \ac{USL} of t1 is more sensitive than the \ac{USL} of t2.
The fault classification based on t2 had an excellent recall of 1.00 and a precision of 0.64. Indeed, as can be seen in the bottom 
panel of Fig.~\ref{fig:PLS_productionCoils_t2}, the fault classifier did not miss any faults and it classified some samples that were close to the \ac{USL} as faults. 
The corresponding $F_3$ score was high: 0.95.

\begin{figure}
\centering
\includegraphics[width=0.95\linewidth]{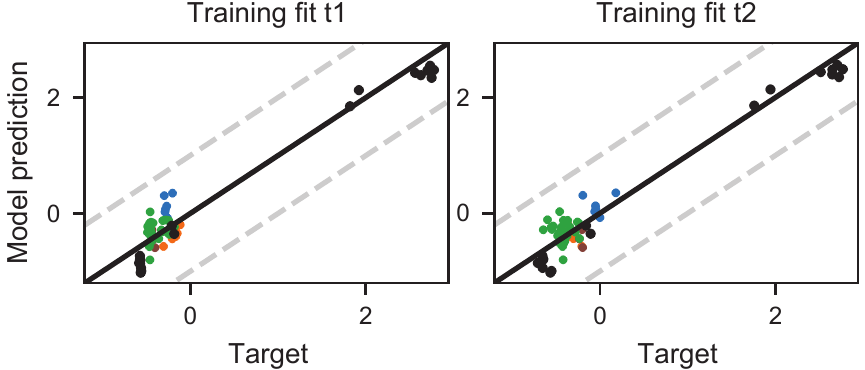}
\caption{Training fit (model vs. target) of \ac{PLS} with number of \ac{PLS} components $k=1$ on all production samples. \textit{Left}: model vs. target for material property t1. \textit{Right}: model vs. target for material property t2.}
\label{fig:PLS_trainingfit}
\end{figure}
Fig.~\ref{fig:PLS_trainingfit} shows the training fit of the \ac{PLS} model when all data was included in the training set, which is the same fit of which the loadings are shown in Fig.~\ref{fig:PLS_loadingplot}. 
It can be seen that the linear model had a sufficient complexity to fit the data. 
As seen from the cross-validation results in 
Fig.~\ref{fig:PLS_productionCoils_t2}, the points from the testcoil had a large influence on the model fit. 
We assume in the rest of the discussion that the weaker linear relationship observed when excluding the testcoil points is related to the fact that these points were only in a small value range and had additional noise caused by the distance between the tensile test and the sensor measurement. It is assumed that the linear relationship as observed for the entire dataset generalizes, see the discussion in Sec.~\ref{sec:Discussion}.

\subsection{Relation of material properties to known production faults}
Besides predicting if the material is out of specification bounds based on non-invasive measurements we are interested whether such measures can be related to the occurrences of product faults
recorded during production.
The \ac{PLS} model fitted on all available data points was used to estimate material properties from the
sensor measurements taken during production. Subsequently, the estimations were compared to the logged faults. As an example case of what can be encountered in production, we first show the result of the model on the known suspicious testcoil and then consider the other logged faults from the rest of production.

\subsubsection{Testcoil results}
\begin{figure}
\centerline{\includegraphics[width=0.95\linewidth]{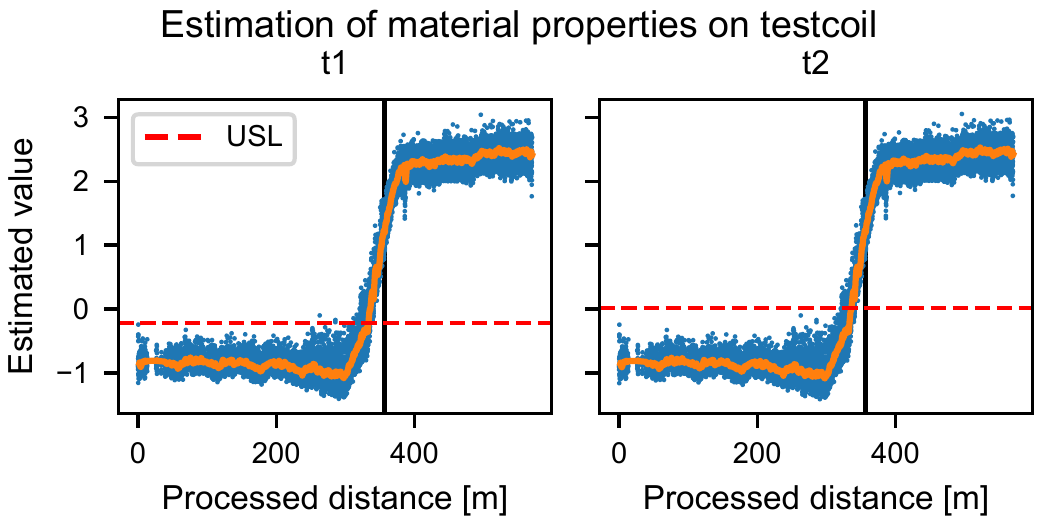}}
\caption{Estimation of the material properties t1 (\textit{left}) and t2 (\textit{right}) based on the sensor measurements made on the testcoil. \textit{Solid orange line}: moving average over 50 values. \textit{Solid black line}: marks the point at which the related production coil was removed from the production line.}
\label{fig:testrol_estimation}
\end{figure}
In Fig.~\ref{fig:testrol_estimation} the model estimations of the material properties are shown for the test coil. Halfway the coil, the material properties drifted out of specification. The point at which production with the related coil was stopped due to cracks occurring in the press has been marked in the figure. As can be seen, this is right after the material properties exceeded the specifications.

\subsubsection{Production data}
\begin{figure}
\centering
\includegraphics[width=0.95\linewidth]{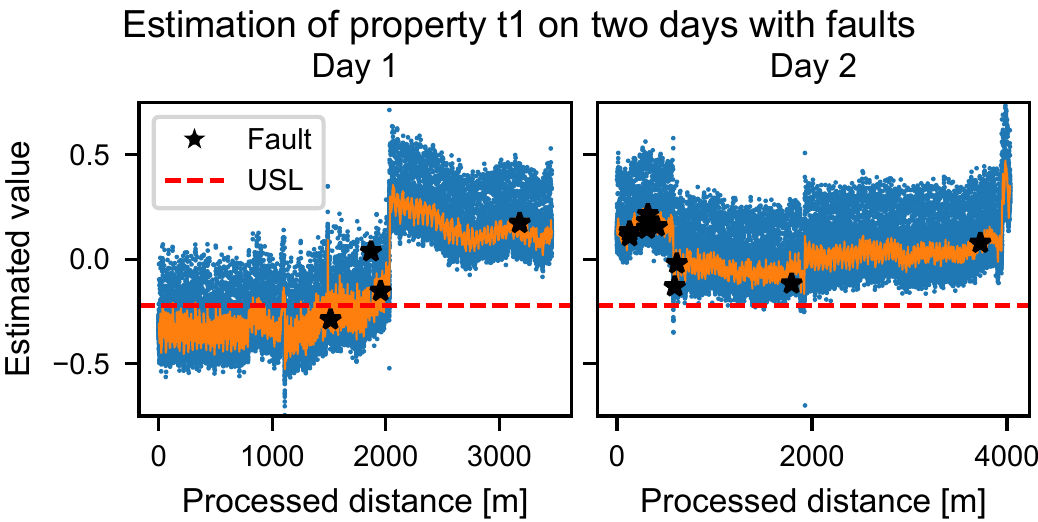}
\caption{Model estimation of material property t1 for two full production days. \textit{Black stars} indicate the model predictions made using the sensor measurements that were linked to product faults. \textit{Solid orange line}: moving average over 50 values.}\label{fig:t1_predictions_labeled_faults}
\end{figure}

In total we got 17 measurement identifiers of strip steel that were linked to faults later in production and for the rest of the dataset we got hours at which faults occurred. Of the 17 measurements, there were 12 predictions by the \ac{PLS} model that exceeded the \ac{USL} of t1 and t2.
Four predictions only exceeded the \ac{USL} of t1 and the remaining one was within the specifications.
However, a large fraction of the estimated material properties in these coils were out of specification but not labeled as faults in production
as is shown in Fig.~\ref{fig:t1_predictions_labeled_faults}.
Therefore the question arises if measurements from predicted material faults that were not connected to reported product faults could be distinguished from those related to faults. 
If this was true a classifier that works on small sample sizes should distinguish those cases. 
In order to test this hypothesis we trained the supervised \ac{GMLVQ} \cite{Schneider2009} model using the implementation from \cite{vanVeen2021} on the 16 labeled sensor measurements (positive class) and 16 randomly chosen measurements that did not cause faults but also had estimated out of specification material properties (negative class). 
Out of 100 random cross-validation splits with 8 samples validation set size and training with early stopping, the mean validation area under the ROC curve
was $0.58$, which is barely above random indicating that it could not be distinguished well.
This suggests that the prediction of undesired material properties does not necessarily cause a fault every time, 
but rather increases the risk of a production fault.

\begin{figure}
\centering
\includegraphics{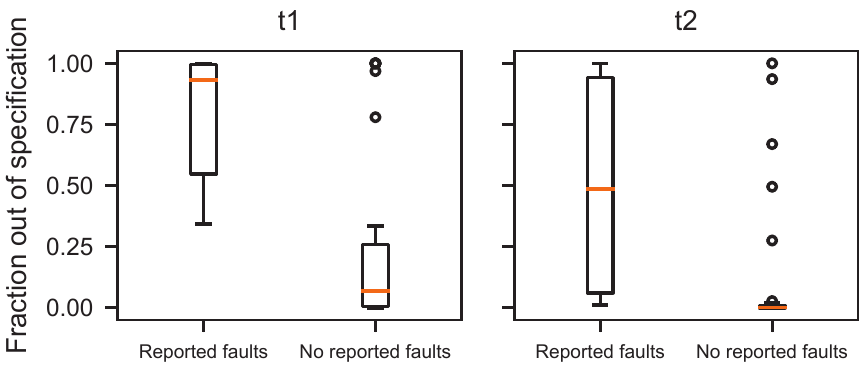}
\caption{Fraction of out of specification model estimations of the material properties t1 and t2 for coils with reported faults and without reported faults.}
\label{fig:t1t2_fractionOutSpec_faults}
\end{figure}
As indication of the risk for a fault we computed the fraction of estimated material properties that were out of specification for each production coil with at least 2000 measurements (40 coils). 
Fig.~\ref{fig:t1t2_fractionOutSpec_faults} shows that for the six coils with reported faults, the fraction of estimated out of specification material properties was significantly higher than for the 34 coils without reported faults. 
Especially for t1, the great majority of production coils without reported faults had a lower fraction of out of specification t1 than the coils with reported faults.

\section{Discussion} \label{sec:Discussion}
From the cross-validated \ac{PLS} performance, we found evidence that the relevant information about the material properties was mainly contained in the higher frequency sensor variables. 
The latent variables of the sensor measurements and the targets were 
linearly correlated. 
We demonstrated that the sensor variables had different levels of measurement noise and that using linear regression with one of the least noisy variables resulted in similar estimation performance as the \ac{PLS} model.
Hence, the results are robust with a comparably wide range of frequencies of the sensor.

The model fitting was heavily influenced by the suspicious test coil measurements which covered a significantly larger variety of material properties than the other coils. 
However, coils with material properties close to the \ac{USL} also conformed to approximately the same linear relationship. 
Given that the production coils fell in a small range of material properties, it is important that such measurements are performed as accurately as possible.
We confirmed that in a number of cases the sensor measurements showed considerable variations, hence the distance between the tensile test and the sensor measurements for the production coils added uncertainty to the true value of the measurement at the location of the tensile test.
Moreover, the time administration of new coils was not always exact, such that in a few cases the closest sensor measurements to the location of the tensile test could not be determined and the averaging was done over a suboptimal sample. 
The accuracy of the the current \ac{PLS} model can be 
verified by additional
tensile test samples from the coils in production and comparing the estimation with the tensile test result.

In the cross-validation the estimations of material property t2 were slightly better than the estimations of t1. Likewise, the material fault classification based on thresholding with respect to the \ac{USL} had a much better recall for t2 than for t1, which is a crucial performance indicator in mass production settings.
However, the results showed that the \ac{USL} of t2 is less sensitive than the \ac{USL} of t1.
Therefore, when relating the material specification predictions to actual reported faults during production, the fraction of violations of the \ac{USL} of t1 was always large for the coils with reported faults. 
In scenarios with a clear drift in material properties, such as the one of the testcoil, the estimation of material properties from the inline \ac{NDT} measurements can prevent material that is far out of specification from entering the production line in the future. 
In these situations the insufficient material quality is most likely the culprit causing production faults. 
In more subtle scenarios, where the estimated material properties were just above the \ac{USL}, the production of the great majority of products did not result in reported faults.
Hence, in order to prevent faults in these situations, it may be crucial to estimate a risk value of faults given the sensor measurements and raise an alert
or adjust the parameters of the production machinery
suitable for the encountered material.

\section{Conclusion and Outlook} \label{sec:Conclusion}
This contribution discusses an exemplary industry 4.0 case: the real-time fault detection and quality control in a mass production line.
Material measurements gathered by an \ac{NDT} soft sensor were analysed in three scenarios:
firstly, measurements taken on deliberately altered material showed that these modifications can be detected by the sensor.
Secondly, a \ac{PLS} model was fitted and validated on measurements taken from several coils in production,
after which it was used to estimate material properties of a suspicious twin to a coil that had to be removed from production and evaluated with destructive testing.
And lastly an analysis of 108 km of coil 
encountered during the full run of this experiment with reported production faults.
We showed the potential of the strategy 
in preventing insufficient material quality from entering the production line. 
In the future, the prevention of these faults could save extremely high costs due to machinery damage. 
Furthermore, the material specification may not always directly lead to faults, but could have a direct influence on the durability of tooling.
We also demonstrated 
evidence 
in preventing the more subtle faults, by revealing the relationship between large fraction of out of specification estimations and reported faults. 
A future direction is to combine the model estimations and risk determination with 
machine parameters, to identify optimal settings for the specific properties of the material, which has the potential to 
widen the material's specification limits.
Further investigations will incorporate process knowledge, such as the physics of the sensor,
other inline measurements and the interplay of the tooling with certain material properties for the prevention of faults.

\bibliographystyle{IEEEtran}

\end{document}